# INDOOR-LiDAR: Bridging Sim2real for Robot-Centric Indoor Scenes Perception – A Hybrid Point Cloud Dataset

Haichuan Li  
haicli@utu.fi

Panos Trahanias  
trahania@ics.forth.gr

Changda Tian  
dada@ics.forth.gr

Tomi Westerlund  
tovewe@utu.fi

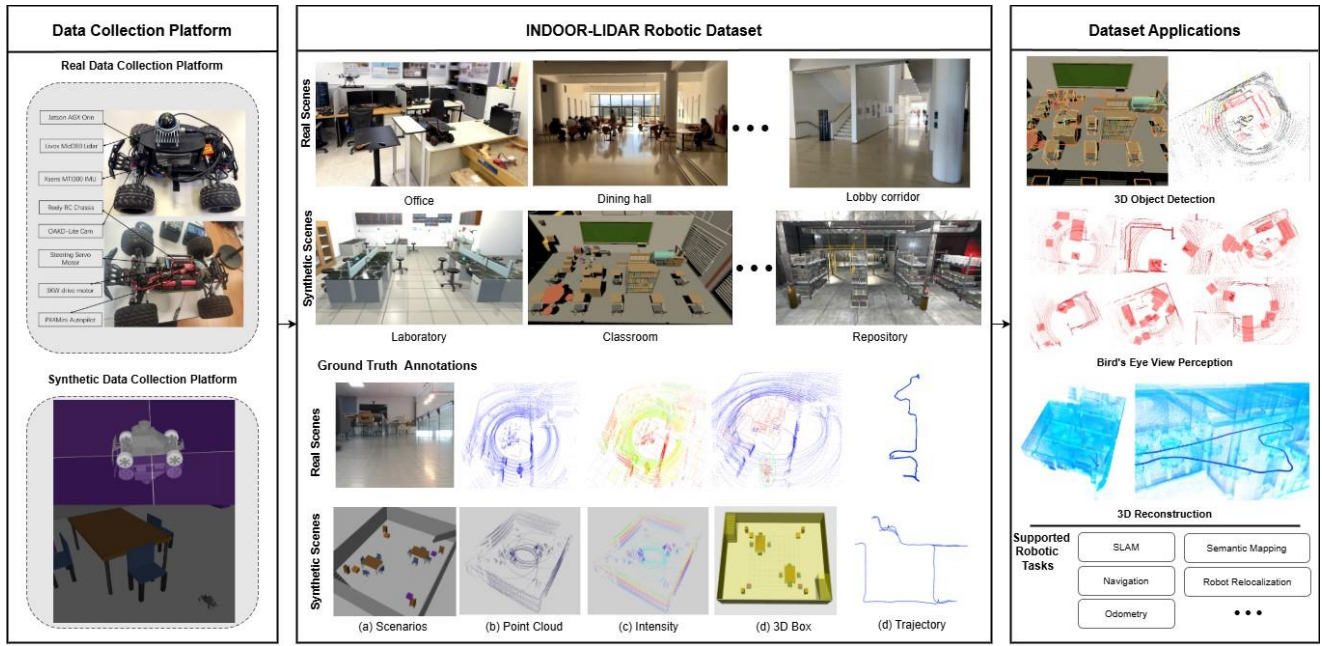

Figure 1. Overview of our INDOOR-LIDAR Dataset framework and applications. The comprehensive data acquisition, processing, and application pipeline encompasses three components: (1) data collection, (2) dataset composition and (3) dataset application. The data collection on the left panel details both real and synthetic data collection platforms, featuring the hardware for physical acquisition, alongside a simulated counterpart for synthetic data generation. The center panel presents dataset samples across diverse indoor scenes, organized as both real and synthetic environments, with corresponding annotations including scenarios, point clouds, intensity maps, 3D bounding boxes, and trajectories. The right panel demonstrates applications of our dataset, including 3D object detection, bird's eye view perception, 3D reconstruction, and various robotic tasks such as SLAM, semantic segmentation, navigation, relocalization, and odometry. Our dataset bridges the gap between real-world and synthetic environments for robust development and evaluation of robotic perception.

## Abstract

*We present INDOOR-LIDAR, a comprehensive hybrid dataset of indoor 3D LiDAR point clouds designed to advance research in robot perception. Existing indoor LiDAR datasets often suffer from limited scale, inconsistent annotation formats, and human-induced variability during data collection. INDOOR-LIDAR addresses these limitations by integrating simulated environments with real-world scans acquired using autonomous ground robots, providing consistent coverage and realistic sensor behavior under controlled variations. Each sample consists of dense point cloud data enriched with intensity measurements and KITTI-style annotations. The annotation schema encompasses common indoor object categories within various scenes. The simulated subset enables flexible configuration of layouts, point densities, and occlusions, while the realworld subset captures authentic sensor noise, clutter, and domain-specific artifacts characteristic of real indoor settings. INDOOR-LIDAR supports a wide range of applications including 3D object detection, bird's-eye-view (BEV) perception, SLAM, semantic scene understanding, and domain adaptation between simulated and real indoor domains. By bridging the gap between synthetic and realworld data, INDOOR-LIDAR establishes a scalable,*

*realistic, and reproducible benchmark for advancing robotic perception in complex indoor environments.*

1. Introduction

In recent years, the research community has benefited from large-scale camera-based indoor datasets such as ScanNet [5], Matterport3D [3], and Replica [40]. These visual datasets have propelled scene reconstruction and semantic understanding, but their reliance on illumination, textures, and material appearance limits their robustness when transferring between environments or between simulation to reality. The gap is especially significant for camera-based systems, where simulated imagery cannot fully replicate real-world lighting, reflections, and textures.

LiDAR has emerged as a key sensing technology for this purpose, providing precise and dense 3D point cloud data that remain reliable under varying lighting conditions. These characteristics make LiDAR indispensable for perception and scenario understanding in indoor robotics [26, 44]. LiDAR, encodes the geometric structure of environments, capturing shapes, dimensions, and spatial relationships that remain consistent across simulated and real scenes. Objects such as tables or shelves preserve their geometry independent of color, texture, or illumination. Moreover, LiDAR sensing remains dependable in environments where illumination is restricted or highly variable, e.g. chemical storage areas, clean rooms, or laboratory facilities, where strong or fluctuating light sources are undesirable or unsafe [2]. This insensitivity to lighting and surface appearance makes LiDAR a uniquely stable modality for domain adaptation and robust 3D perception [34]. However, the advancement of perception algorithms fundamentally depends on the quality of the datasets.

Therefore, LiDAR-based indoor datasets have attracted growing research attention as well. Works such as LiDAR-Net [11] provide extensive real-scanned indoor point clouds captured with Mobile Laser Scanning (MLS), representing realistic sensor noise and spatial variability. Similarly, 3DSES [25] targets the AEC (Architecture, Engineering, and Construction) domain through high-density Terrestrial Laser Scanning (TLS) with CAD alignments, facilitating precise scan-to-BIM research. These datasets have expanded the capabilities of perception systems, especially for semantic segmentation and structural modeling tasks, but still do not fully represent the sensor configuration and perception conditions encountered by mobile robots.

A notable limitation arises from handheld data acquisition. When an operator carries a LiDAR sensor, the human body physically blocks a significant portion of the scanner's field of view, typically obstructing the rear or lower hemisphere. This creates substantial blind zones that prevent 360° environmental perception [51, 52]. As a result, nearly half of the point returns are absent or dominated by reflections from the operator's body, producing incomplete and biased spatial information [16, 43]. Such non-uniform coverage makes handheld data unsuitable for robot perception systems, which require uninterrupted panoramic sensing to ensure safe navigation, obstacle detection, and situational awareness in all directions [13, 19]. These limitations underscore the need for a dataset designed from the robot's operational perspective. In contrast, a robot-mounted LiDAR maintains a clear, unobstructed vantage point and provides continuous 360° coverage of the surroundings, enabling comprehensive environment understanding. In contrast, since our dataset is collected directly from a robot's sensor perspective, the resulting data are intrinsically aligned with real robotic operation, making the trained models and perception algorithms easier to implement and more transferable to real systems without extensive adaptation.

To address these challenges and fully leverage LiDAR's geometric consistency, we introduce INDOOR-LIDAR, a dedicated dataset designed to advance research in robot-centric indoor perception. The primary contributions of our work are threefold:

- Hybrid Real and Simulated Environments: We proposed a comprehensive robot-based indoor LiDAR dataset that combines procedurally generated simulated spaces with authentic real-world scans, enabling systematic investigation into sim-to-real transfer for perception.
- Robot-Centric 360° Data Acquisition: All data are captured from the viewpoint of an autonomous ground robot, guaranteeing complete 360° field-of-view coverage without occlusion and reflecting the real motion and perception dynamics of mobile platforms.
- Comprehensive Detection Benchmarks: In addition to raw data, the dataset provides standardized splits and baseline evaluations using popular perception algorithms, offering strong reference points for future research.

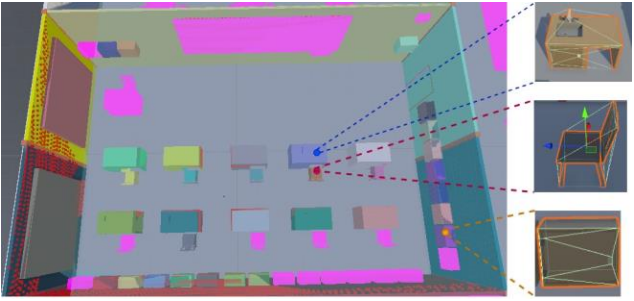

Figure 2. LiDAR-based 3D Detection and Segmentation. Visualization of our pipeline processing a raw indoor LiDAR mesh into detected, segmented, and geometrically modeled objects.

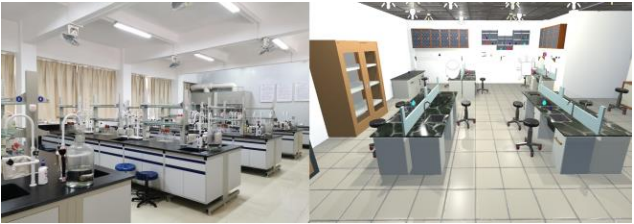

Table 1. Comparison of the characteristics of various point cloud datasets from the literature. Note that INDOOR-LiDAR is the only indoor MLS dataset that includes intensity, point level annotations, 360° field of view and 3D simulated model. The robot-mounted collection method makes the AI models trained from our dataset more smoothly to implement on real devices.

| Name | Environment | Classes | Extent | Points (M) | 360° | Intensity | 3D model | Source |
|---|---|---|---|---|---|---|---|---|
| Oakland [24] | Outdoor | 44 | - | 1.6 | ✗ | ✗ | ✗ | MLS |
| Paris-rue-Madame [35] | Outdoor | 17 | 160 m | 20 | ✗ | ✓ | ✗ | MLS |
| IQmulus [48] | Outdoor | 8 | 10000 m | 12 | ✗ | ✓ | ✗ | MLS |
| Semantic 3D [12] | Outdoor | 8 | - | 4000 | ✓ | ✓ | ✗ | TLS |
| Paris-Lille-3D [33] | Outdoor | 9 | 1940 m | 143.1 | ✓ | ✓ | ✗ | MLS |
| SemanticKITTI | Outdoor | 25 | 39200 m | 4500 | ✓ | ✓ | ✗ | MLS |
| Toronto-3D [42] | Outdoor | 8 | 1000 m | 78.3 | ✓ | ✓ | ✗ | TLS |
| ScanNet++ [55] | Indoor | - | 15000 m² | 20 | ✓ | ✗ | ✗ | TLS |
| LiDAR-Net [11] | Indoor | 24 | 30000 m² | 3600 | ✗ | ✓ | ✗ | MLS |
| 3DSES [25] | Indoor | 18 | 832 m² | 674 | ✗ | ✓ | ✓ | TLS |
| Ours(Sim) | Indoor | 20 | 5000 m² | 1200 | ✓ | ✓ | ✓ | MLS |
| Ours(Real) | Indoor | 15 | 800 m² | 150 | ✓ | ✓ | ✓ | MLS |
| Indoor Modelling (Khoshelham et al., 2017) | Indoor | ✗ | 2824 m² | 127 | ✗ | ✗ | ✓ | 5 sensor |
| Craslab (Abreu et al., 2023) | Indoor | ✗ | 417 m² | 584 | ✗ | ✓ | ✓ | TLS |

Figure 3. The real and simulated scene example, a chemistry lab.

## 2. Related Work

The advancement of 3D perception for robotics is deeply intertwined with the development of high-quality datasets. Such resources are crucial not only for training and validating deep learning models but also for establishing standardized benchmarks that drive the field forward. The rich and cluttered nature of indoor spaces has made them a prime target for 3D scene understanding research. Datasets in this area can be broadly categorized by their primary sensor modality: camera-based or LiDAR.

### 2.1. Camera-based Indoor Datasets

Early and influential large-scale indoor datasets were primarily captured using camera sensors. The NYUv2 [38] and SUN RGB-D [39] datasets provided foundational, richly annotated collections of indoor scenes that fueled initial research into 3D semantic segmentation and object detection. Following these, ScanNet [5] and Matterport3D [3] significantly scaled up data collection by leveraging RGB-D sensors to create thousands of 3D reconstructed edges from diverse indoor spaces. Building on these, the Replica dataset [40] introduced high-fidelity 3D reconstructions of 18 indoor scenes using a custom RGBD rig, providing pseudo point cloud, textures, and semantic annotations. The ARKitScenes dataset [1] further expanded the scope by offering 3D edges, bounding boxes, and semantic labels for robust indoor scene analysis. Similarly, the Habitat-Matterport 3D(HM3D) dataset [31] provides high-quality 3D reconstructions of indoor environments with navigable edges and semantic annotations. Recent advancements have focused on enhanced precision and specialized applications. The Mirror3D dataset [41] augments existing RGB-D datasets (NYUv2, ScanNet, Matterport3D) with annotated mirror instances and 3D planes, addressing challenges in reconstructing reflective surfaces for improved segmentation and 3D modeling. ScanNet++ [55] extends ScanNet with 460 high-fidelity scenes captured using DSLR cameras and laser scanners, achieving submillimeter accuracy and providing detailed semantic annotations for tasks like novel view synthesis and fine-grained scene understanding. The StructScan3D v1 dataset [30] offers 2,594 RGB-D frames from indoor buildings captured with Kinect Azure, annotated for six structural elements to support semantic segmentation and Building Information Modeling (BIM). Additionally, IAM Benchmarks [15] introduces three new RGB-D segmentation benchmarks, integrating visual and depth cues for fine-grained object recognition. The IL3D dataset [59] provides over 10,000 indoor scenes with semantic and spatial annotations derived from RGB-D scans, enabling 3D arrangement generation and scene understanding.

2.2. LiDAR-based Indoor Datasets

To overcome the limitations of RGB-D, several recent datasets have turned to LiDAR technology. LiDAR-

Separately, 3DSES [25] utilized high-precision Terrestrial Laser Scanners (TLS) to create dense, survey-grade point clouds for scan-to-BIM applications, uniquely pairing the scans with 3D CAD models. In addition to these, ConSLAM [46] provides a periodically collected real-world construction dataset using LiDAR scans, supporting SLAM and progress monitoring in dynamic indoor environments. Hilti-Oxford [56] offers a millimeter-accurate multi-sensor benchmark with LiDAR for SLAM, focusing on indoor construction sites. FusionPortableV2 [50] introduces a unified multi-sensor dataset including indoor building and lab sequences captured via LiDAR, aimed at generalized SLAM across diverse platforms. 3DRef [58] presents an indoor LiDAR dataset with annotations for reflection detection, enhancing robustness in point cloud processing for 3D reconstruction. FIORD [10] delivers a fisheye indooroutdoor dataset with LiDAR ground truth,

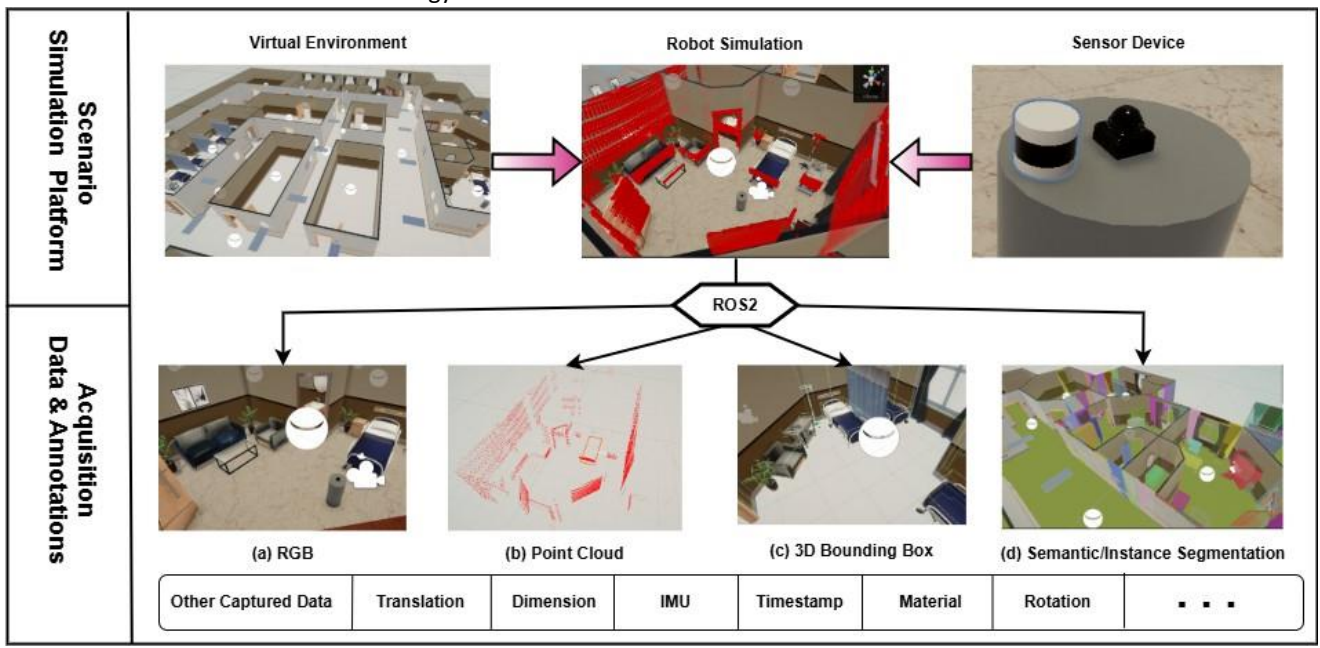

Figure 4. Integrated robotic simulation and data acquisition framework. The diagram illustrates our comprehensive pipeline for robotic perception data generation and processing. The upper section presents the Scenario Simulation Platform comprising three key components: (1) a Virtual Environment showing the 3D model of an indoor space; (2) Robot Simulation displaying the simulated robot in a rendered environment with obstacle detection; and (3) Sensor Devices. The lower section demonstrates the Data Acquisition & Annotations pipeline powered by ROS2, producing four data modalities: (a) RGB imagery of the simulated environment; (b) Point Cloud representations capturing spatial information; (c) 3D Bounding Box object detection; and (d) Semantic/Instance Segmentation with colorcoded object classification. Additional metadata including translation, dimension, IMU readings, timestamps, material properties, and rotation measurements supplement the core datasets.

Net [11] presented a massive-scale indoor dataset captured via a handheld Mobile Laser Scanning (MLS) system.

specifically tailored for 3D scene reconstruction applications. In addition to these, IILABS 3D [32] offers a multi-LiDAR indoor SLAM benchmark with synchronized data from various 3D and 2D LiDAR sensors, IMU, and wheel odometry, including high-precision ground truth for evaluating SLAM in industrial lab settings. Indoor Multi-Modal MultiFloor Dataset [17] presents a challenging multi-sensor indoor dataset spanning multiple building floors, featuring LiDAR alongside cameras and IMU to test SLAM robustness in scenarios like perceptual aliasing and visual degradation. Hilti-Oxford [56] provides a millimeter-accurate multi-sensor benchmark with LiDAR for SLAM, emphasizing indoor construction sites with dynamic elements. ConSLAM [46] introduces a periodically collected indoor construction dataset using LiDAR scans, aiding SLAM and progress monitoring in dynamic environments. FusionPortableV2 [50] includes unified multi-sensor indoor sequences captured via LiDAR on various platforms, targeted at generalized SLAM. LiFall [49] delivers a LiDAR-based dataset for elderly fall detection in indoor environments, with simulated fall scenes for neural network training.

## 2.3. Simulation and Domain Adaptation

Modern simulation environments, including NVIDIA Isaac Sim [23], Unreal Engine [6], Unity [47], Gazebo [18], and MuJoCo [45], have become indispensable components of robotics research. These platforms facilitate the safe, scalable, and cost-effective generation of synthetic training data, enabling controlled experimentation across diverse conditions. Nevertheless, their effectiveness is frequently limited by the well-known *sim-to-real* domain gap, which arises from discrepancies between simulated and real-world sensory observations [57]. In this context, LiDAR sensing offers a significant advantage. Unlike cameras, LiDAR measurements primarily encode geometric information [34]. The spatial structure, shape, and dimensions of objects exhibit far greater consistency between virtual and physical environments than their visual features. This property renders LiDAR an appealing modality for research in domain adaptation and hybrid dataset development, where geometric fidelity is crucial for transferring perception models from simulation to the real world [7, 53].

## 2.4. Benchmarks for Perception Tasks

Progress in 3D scene understanding has relied heavily on standardized datasets that catalyze algorithmic innovation across multiple perception tasks. In outdoor robotics, the KITTI benchmark [8] established a foundation for evaluating 3D object detection, tracking, and localization, defining data formats and metrics that remain influential. However, models trained on outdoor datasets such as PointNet [27], PointNet++ [28], and VoteNet [29] often exhibit domainspecific limitations when applied to indoor robotics. Indoor environments pose unique challenges, including high clutter, short sensing range, occlusions, and dense point clouds, demanding tailored benchmarks and evaluation protocols.

To address this need, the INDOOR-LiDAR dataset introduces a unified benchmark suite for *scenario understanding in indoor robotics*. This suite encompasses three complementary tasks: (1) 3D Object Detection: We provide evaluation protocols compatible with KITTI-style annotations while focusing on indoor-scale perception. (2) Bird'sEye-View (BEV) Perception: Bird's-eye-view understanding offering a top-down perspective is critical for spatial reasoning and navigation in robotics, especially for ground robots operating in cluttered indoor spaces. (3) SLAM: Simultaneous Localization and Mapping (SLAM) is a foundational capability for autonomous indoor robots, enabling them to build maps of unknown environments while concurrently localizing themselves within those maps.

## 3. Methodology

### 3.1. Simulated Data Generation

The synthetic track of INDOOR-LiDAR is generated using a custom simulation pipeline that couples Unity and MuJoCo for environment modeling with Taichi for massively parallel LiDAR ray tracing. This design enables scalable scene diversity, physically grounded interactions, and efficient generation of dense point clouds with perfect groundtruth annotations.

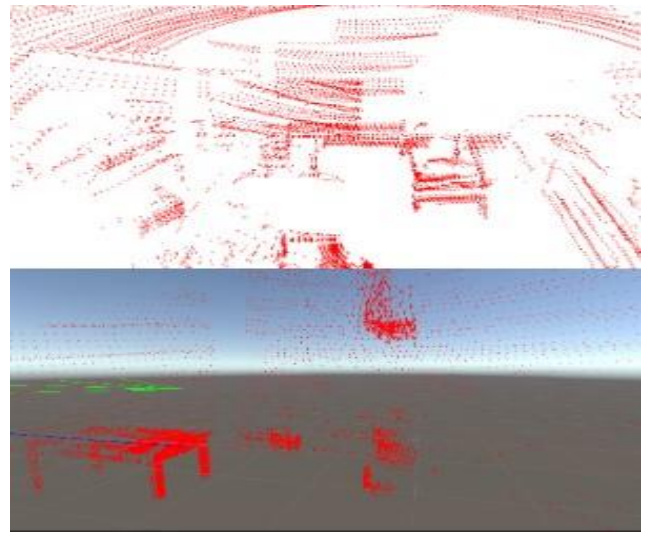

Figure 6. LiDAR-based 3D Detection and Segmentation. Visualization of our pipeline processing a raw indoor LiDAR mesh into detected, segmented, and geometrically modeled objects.

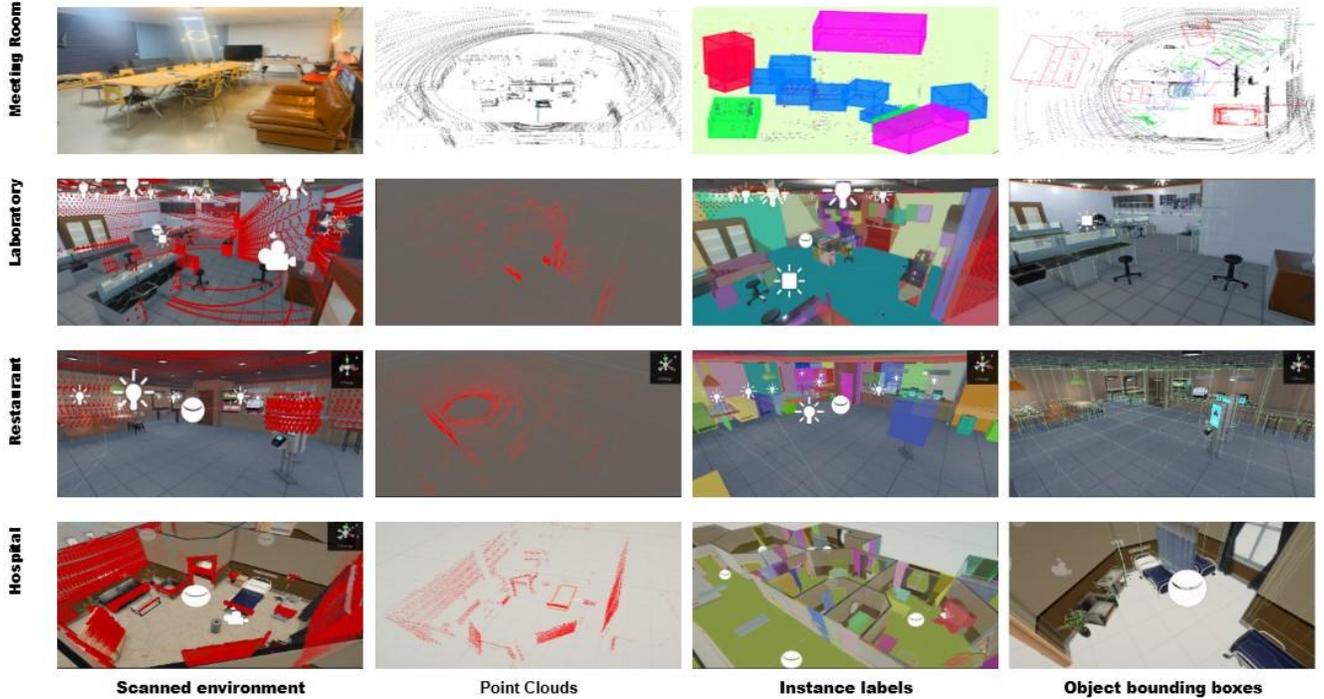

Figure 5. Multimodal data representation comparison between real-world and simulated environments in the INDOOR-LIDAR dataset. The matrix is organized by environment type (rows) and data representation modality (columns). The first row shows a real-world meeting room, while the remaining rows display simulated environments (laboratory, restaurant, and hospital). Each environment is represented in four formats: scanned environment (original view), point clouds (spatial distribution of LiDAR returns), instance labels (semantic segmentation with color-coded object categories), and object bounding boxes (3D localization). This comprehensive visualization demonstrates the substantial differences in data characteristics between real and simulated captures, particularly in point cloud density, noise patterns, and boundary precision. The dataset's paired real-simulated structure facilitates quantitative evaluation of perception algorithms across the sim-to-real gap and enables the development of robust transfer learning techniques for indoor robotic applications..

Procedural Indoor Scene Generation: We build each virtual environment in Unity and Mujoco using procedurally sampled floor plans populated with common indoor structures and objects. Object positions, orientations, densities, and room layouts are adjustable per scene, enabling broad coverage of indoor spatial configurations.

Taichi LiDAR Simulation: To generate LiDAR scans, we integrate a high-performance ray-casting module implemented in Taichi [14]. Given a sensor pose $\mathbf{T}_s \in SE(3)$, a dense set of rays is emitted using spherical sampling:

$$\mathbf{d}(\theta, \phi) = \begin{bmatrix} \cos\phi\cos\theta, & \cos\phi\sin\theta, & \sin\phi \end{bmatrix}^\top, \quad \mathbf{p}(t) = \mathbf{o} + t\,\mathbf{d}.$$

where $\mathbf{o}$ is the sensor origin. Each ray is transformed into the local frame of each object, and analytic intersection routines (per geometry type) are used to compute candidate hit distances. The nearest positive solution yields the final return, producing a point cloud $\mathcal{P} = \{\mathbf{p}_i\}_{i=1}^{N}$ in the LiDAR frame. The Taichi implementation parallelizes ray evaluation across all rays and objects, with kernel fusion and preallocated memory structures to minimize overhead. As a result, the simulator maintains near real-time performance for scans exceeding 100k rays per frame and supports dynamic object motion or scene changes without reinitialization.

Synthetic Ground-Truth Extraction: Because all geometry and object poses are known, the simulator provides perfect ground-truth annotations. For each frame, 3D bounding boxes, object classes, and poses are queried directly from the virtual world state. And only the scanned objects' bounding boxes will be recorded.

Robotic Platform and Sensor Setup: Real-world LiDAR data were collected using a custom unmanned ground vehicle (see Fig. 1). The platform employs an Ackerman RC chassis with stable low-vibration motion characteristics and is designed specifically for navigation in confined indoor environments. A LiDAR sensor is rigidly mounted at the top of the chassis, providing an unobstructed 360° field of view at a height representative of typical service and inspection robots. An OAKD-Lite camera is mounted in the front of the UGV. An Xsens Mti300 IMU is mounted on the chassis. A

Jetson AGX Orin computation module acts as the center of sensor fusion and control. The UGV can be manually teleoperated and autonomously driven through a variety of indoor spaces-including offices, laboratories, hallways, and cluttered utility areas—to capture realistic sensor observations under diverse geometric and environmental conditions. The resulting scans reflect the true perception profile of a ground robot operating in everyday indoor settings.

Annotation Process: Ground-truth annotations for the real-world scans were produced using a meticulous semi-automatic pipeline. The open-source annotation tool SUSTechPOINTS [20] was used as the primary interface for this process. Annotators first performed an initial labeling of object instances, after which algorithmic aids helped refine the bounding box parameters. Every single label was then manually verified by an expert annotator to ensure high accuracy and consistency across the dataset. The annotations include object classes, position and dimension of 3D objects, and corresponding orientation, which are then converted to the final KITTI-style format.

Dataset Organization and Format: Each frame in the dataset consists of a point cloud file and a corresponding label file, structured as follows:

- Input Point Clouds: Lidar scans are stored in .bin files. Each file contains an $N \times 4$ NumPy array, where $N$ is the number of points in the scan, and the four columns represent the point's 3D coordinates and its intensity value (x, y, z, intensity).
- Labels: Annotations are provided following the KITTI label2 format. Each line in the file corresponds to an object instance and records its properties, including category, 3D dimensions, 3D location, and yaw angle ($r_y$).
- IMU Data: For real-world scans, synchronized IMU data is provided in separate text files, containing timestamps and orientation information to facilitate sensor fusion tasks and SLAM applications.
- Timestamps: Each LiDAR frame carries a ROS 2 timestamp (nanosecond resolution), which serves as a global time reference for synchronizing IMU packets and robot odometry. This ensures consistent multi-sensor alignment for mapping, state estimation, and SLAM evaluation.

## 4. Experiments

To validate the INDOOR-LiDAR and establish baselines for future work, we conduct comprehensive experiments using a variety of popular models. The primary goal is to provide a standardized benchmark that evaluates performance on both the simulated and real-world tracks of our dataset, highlighting challenges in indoor robotic perception.

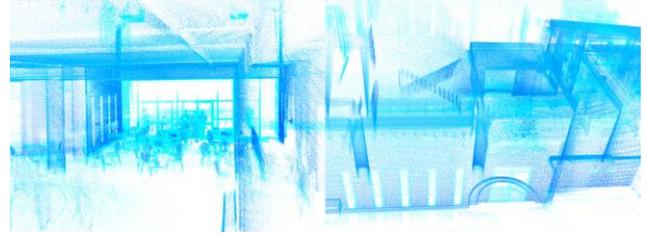

Figure 7. Representative 3D reconstructions from diverse indoor environments included in INDOOR-LiDAR. The subfigures depict a cafe scene and a stairwell from the real-world collection.´

### 4.1. Evaluation

To provide a multifarious evaluation of model performance, we report metrics across three aspects of the detection task: • Classification Performance: We use Precision (P) to evaluate a model's ability to correctly classify detection.
• Bounding Box Quality: We evaluate the geometric accuracy of the predicted bounding boxes using several metrics: Mean IoU across all true positives, Accuracy at different IoU thresholds, and the L1 and L2 distance errors. For SLAM applications, we choose 2 efficient algorithms to evaluate localization and mapping performance: DLIO [4] and LIO-SAM [36]. The results are shown in Appendix. 9.

### 4.2. Baseline Models

We evaluate a representative set of BEV-based and full 3D detectors. BEV Detectors: We include PillarNet [9], PointNet++ (BEV projection) [28], BEV MAE [21], and GroupFree3D-BEV [22]. 3D Detectors: For direct 3D reasoning, we benchmark voxel-based methods (VoxelNet [60], SECOND [54]), point-based methods (PointRCNN [37], VoteNet [29]), and transformer-based models (GroupFree3D [22]). Further architectural and training details are provided in the Supplementary.

Table 2. Classification performance and bounding box metrics on INDOOR-LiDAR simulated Test Set. M1=PillarNet, M2=PointNet++, M3=BEV MAE, M4=GroupFree3D

| Metric | M1 | M2 | M3 | M4 |
|---|---|---|---|---|
| Classification Precision (P) | | | | |
| Table | 0.47 | 0.45 | 0.50 | 0.55 |
| Chair | 0.22 | 0.18 | 0.19 | 0.20 |
| Shelf | 0.05 | 0.05 | 0.10 | 0.15 |
| Box | 0.05 | 0.08 | 0.07 | 0.06 |
| Stair | 0.34 | 0.62 | 0.56 | 0.50 |
| Bounding Box | | | | |
| Mean IoU | 0.66 | 0.70 | 0.79 | 0.68 |
| Acc@IoU0.25 | 0.89 | 0.91 | 0.92 | 0.90 |
| Acc@IoU0.50 | 0.74 | 0.81 | 0.87 | 0.79 |
| Acc@IoU0.75 | 0.51 | 0.58 | 0.72 | 0.56 |
| L1 | 3.04 | 2.83 | 2.37 | 2.79 |
| L2 | 66.29 | 71.50 | 66.14 | 63.46 |

### 4.3. Benchmark Results and Analysis

We present the benchmark results for the simulated and real-world test sets to clearly analyze model performance.

**BEV Performance on Simulated Data** As shown in Table 2, we observe varying strengths among the four models on our simulated Test Set. For classification precision, no single model dominates across all object categories. GroupFree3D (M4) achieves the highest precision for tables (0.55) and shelves (0.15), while PointNet++ (M2) excels at detecting stairs (0.62) and boxes (0.08). PillarNet (M1) performs best for chair detection (0.22). These results highlight the complementary strengths of different architectures when handling diverse object geometries. For bounding box quality metrics, BEV MAE (M3) demonstrates superior performance across most measures. It achieves the highest Mean IoU (0.79) and consistently outperforms other models at all IoU thresholds (0.92, 0.87, and 0.72 at IoU thresholds of 0.25, 0.50, and 0.75 respectively). M3 also exhibits the lowest L1 error (2.37), although GroupFree3D (M4) achieves the best L2 error (63.46).

**BEV Performance on Real-World Data** Table 3 reveals a notable performance shift when evaluating models on real-world data. For classification, BEV MAE (M3) and GroupFree3D (M4) clearly dominate, achieving substantially higher precision across all major object categories (0.52-0.85), while all models fail on less common objects. Conversely, for geometric accuracy, simpler architectures excel, with PillarNet (M1) achieving the best Mean IoU (0.81) and lowest error rates, followed by PointNet++ (M2) with strongest performance at higher IoU thresholds.

**3D Performance on Simulated Data** Table 4 presents a comprehensive comparison of five state-of-the-art 3D object detection models on simulated data. GroupFree3D (GF) demonstrates superior classification performance, consistently achieving the highest precision across all object categories (0.70-0.85 for common objects like tables, chairs, and stairs; 0.40-0.65 for more challenging categories like boxes and shelves). However, for bounding box metrics, PointRCNN (PR) emerges as the clear leader, with the highest Mean IoU (0.75), best accuracy at all IoU thresholds (0.94, 0.85, and 0.65 at IoU thresholds of 0.25, 0.50, and 0.75 respectively), and lowest error rates (L1: 1.30, L2: 25.80). This indicates that while GroupFree3D excels at identifying object types, PointRCNN achieves more precise geometric localization. VoxelNet and SECOND consistently underperform across all metrics, suggesting their architectures are less suited for indoor object detection tasks compared to the more advanced point-based methods.

Table 3. BEV Classification performance and bounding box metrics on INDOOR-LiDAR Real-World Test Set. M1=PillarNet, M2=PointNet++, M3=BEV MAE, M4=GroupFree3D

| Metric | M1 | M2 | M3 | M4 |
|---|---|---|---|---|
| Classification Precision (P) | | | | |
| Couch | 0.44 | 0.62 | 0.85 | 0.84 |
| Table | 0.27 | 0.25 | 0.56 | 0.55 |
| Person | 0.17 | 0.12 | 0.52 | 0.52 |
| Chair | 0.16 | 0.13 | 0.59 | 0.58 |
| All other | 0.00 | 0.00 | 0.00 | 0.00 |
| Bounding Box | | | | |
| Mean IoU | 0.81 | 0.80 | 0.69 | 0.76 |
| Acc@IoU0.25 | 0.99 | 0.99 | 0.94 | 0.97 |
| Acc@IoU0.50 | 0.89 | 0.90 | 0.79 | 0.85 |
| Acc@IoU0.75 | 0.70 | 0.72 | 0.52 | 0.62 |
| L1 | 1.28 | 1.37 | 2.16 | 1.79 |
| L2 | 12.58 | 15.25 | 29.30 | 22.37 |

Table 4. 3D classification and bounding box performance on simulated data. VN=VoxelNet, SC=SECOND, PR=PointRCNN, VT=VoteNet, GF=GroupFree3D

| Metric | VN | SC | PR | VT | GF |
|---|---|---|---|---|---|
| Classification Precision (P) | | | | | |
| Table | 0.60 | 0.62 | 0.82 | 0.75 | 0.85 |
| Chair | 0.35 | 0.40 | 0.65 | 0.60 | 0.70 |
| Shelf | 0.25 | 0.30 | 0.55 | 0.50 | 0.65 |
| Box | 0.15 | 0.18 | 0.30 | 0.25 | 0.40 |
| Stair | 0.50 | 0.55 | 0.75 | 0.70 | 0.80 |
| Bounding Box | | | | | |
| Mean IoU | 0.55 | 0.58 | 0.75 | 0.68 | 0.72 |
| Acc@IoU0.25 | 0.70 | 0.74 | 0.94 | 0.88 | 0.92 |
| Acc@IoU0.50 | 0.58 | 0.62 | 0.85 | 0.75 | 0.80 |
| Acc@IoU0.75 | 0.30 | 0.34 | 0.65 | 0.50 | 0.55 |
| L1 Loss | 2.10 | 1.95 | 1.30 | 1.65 | 1.50 |
| L2 Loss | 45.50 | 42.10 | 25.80 | 35.20 | 30.15 |

**3D Performance on Real-World Data** Table 5 reveals significant performance shifts when evaluating models on real-world data. Unlike simulated results, VoteNet (VT) emerges as the superior classifier, achieving the highest

| Metric | VN | SC | PR | VT | GF |
|---|---|---|---|---|---|

precision across all categories (0.32-0.60). This represents a notable change from simulated data where GroupFree3D dominated classification tasks. Meanwhile, PointRCNN (PR) maintains its geometric accuracy advantage with the highest Mean IoU (0.78) and best performance across all accuracy thresholds (0.95, 0.88, and 0.68 at IoU thresholds of 0.25, 0.50, and 0.75) and lowest error rates (L1: 1.25, L2: 18.90). Most notably, GroupFree3D (GF), which excelled in simulated environments, shows substantial performance degradation in real-world scenarios, particularly in bounding box metrics where it falls below even the simpler VoxelNet and SECOND models.

Table 5. 3D classification and bounding box performance on real-world data. VN=VoxelNet, SC=SECOND, PR=PointRCNN, VT=VoteNet, GF=GroupFree3D

| | Classification Precision (P) | | | | |
|---|---|---|---|---|---|
| Couch | 0.35 | 0.40 | 0.55 | 0.60 | 0.50 |
| Table | 0.25 | 0.30 | 0.45 | 0.50 | 0.40 |
| Person | 0.20 | 0.22 | 0.35 | 0.38 | 0.30 |
| Chair | 0.15 | 0.18 | 0.30 | 0.32 | 0.25 |
| | Bounding Box | | | | |
| Mean IoU | 0.65 | 0.68 | 0.78 | 0.72 | 0.60 |
| Acc@IoU0.25 | 0.80 | 0.84 | 0.95 | 0.90 | 0.75 |
| Acc@IoU0.50 | 0.65 | 0.70 | 0.88 | 0.80 | 0.60 |
| Acc@IoU0.75 | 0.40 | 0.45 | 0.68 | 0.55 | 0.35 |
| L1 Loss | 1.95 | 1.80 | 1.25 | 1.50 | 2.20 |
| L2 Loss | 35.10 | 32.50 | 18.90 | 24.60 | 40.50 |

## 5. Conclusion

In this paper, we have introduced INDOOR-LiDAR, a novel, large-scale dataset designed to bridge a critical gap in 3D perception for indoor robotics. Our core contribution is a hybrid dataset that combines an extensive, procedurally generated simulated environment with authentic real-world scans captured by a ground robot using universal, roboticsgrade LiDAR sensors. By providing meticulous KITTIstyle annotations for both domains and establishing predefined data splits, we offer a comprehensive and accessible resource for the research community. Through our extensive benchmark experiments on both Bird's-Eye-View and full 3D object detection, we have not only provided a robust performance baseline but also demonstrated the significant challenge of the sim-to-real domain gap in LiDAR perception. Ultimately, INDOOR-LiDAR is more than a collection of data; it is a foundational tool designed to democratize research and accelerate progress in autonomous indoor navigation. By providing data that is both perspectively and economically

# INDOOR-LiDAR: Bridging Sim2real for Robot-Centric Indoor Scenes Perception – A Hybrid Point Cloud Dataset

## Supplementary Material

representative of real-world robotic systems, we hope to spur the development of the indoor perception algorithms that are not only accurate but also robust, reliable, and capable of being deployed in practice.

Acknowledgments : This work was granted by the European Commission's Marie Skłodowska-Curie Action (MSCA) Project RAICAM (GA101072634).

## 6. Example Structure of Simulation Scenes

The INDOOR-LIDAR dataset encompasses a diverse range of indoor environments designed to challenge and evaluate 3D perception algorithms. Fig. 8 illustrates four representative synthetic environments from our dataset, each showcasing different architectural layouts and object configurations. These environments vary significantly in structural complexity, object density, and spatial arrangement, enabling thorough evaluation of perception algorithms across different indoor scenarios. Each scene is meticulously modeled with accurate physical dimensions, material properties, and lighting conditions to closely approximate their real-world counterparts. The white cylindrical markers visible in several scenes represent LiDAR sensor positions used for systematic data collection throughout the environments.

## 7. Example Structure of Real-world Scenes

The visualization in Fig. 9 displays a top-down view of a complex building structure with interconnected hallways, rooms, and open areas rendered in magenta. This point cloud demonstrates the characteristic sparsity and noise patterns inherent in real-world LiDAR data acquisition, contrasting with the clean synthetic environments shown previously. The structural layout includes a main corridor running vertically through the center with branching hallways and adjacent rooms of various sizes. Such real-world scans provide essential validation data for testing algorithms trained on synthetic environments, enabling quantitative assessment of the sim-to-real gap in indoor perception tasks.

## 8. Examples of Scenes' Objects

Our simulation environment is populated with a diverse library of 3D object assets designed to create realistic and cluttered indoor scenes. As shown in the figure, this collection includes a wide range of common household items, such as large furniture (e.g., Bed, Sofa, Table, Stairs), storage units (Cabinet, Shelf), and kitchen appliances (Oven, Microwave oven, Dishwasher, Sink). This variety of object classes, sizes, and geometries is crucial for rigorously training and evaluating our robotic perception algorithms, particularly for tasks like 3D object detection and semantic segmentation, ensuring our models can generalize from simulation to complex real-world environments.

## 9. Examples of Slam using our dataset

To validate the utility of our dataset's real-world sequences, we evaluated the performance of two very popular SLAM algorithms, DLIO [4] and LIO-SAM [36]. Both methods were run on the same challenging indoor trajectories from our dataset to assess their mapping and localization accuracy. As illustrated in the figure, both algorithms successfully processed the sensor data, generating trajectory estimates that are qualitatively very similar. This side-by-side comparison demonstrates that our dataset provides highquality, synchronized sensor data suitable for benchmarking and developing state-of-the-art SLAM systems.

## 10. Examples of BEV-Based detectors usingour dataset

We further demonstrate the utility of our dataset by training a CNN-based Bird's-Eye View (BEV) object detector on the simulation data. This model is designed to perform simultaneous object classification and bounding box regression from the input point cloud. The training and evaluation results, summarized in our test report, are promising. The detector achieved a mean Average Precision (mAP) of 0.679 across all classes (Fig. 17). For the classification task, the model achieved a macro F1-score of 0.698, with detailed per-class performance and a confusion matrix shown in Fig. 12 and Fig. 14, respectively. The bounding box regression performance was particularly strong, with a Mean IoU of 0.987 and an accuracy of 0.990 at an IoU threshold of 0.75. Detailed analyses of IoU and loss distributions are presented in Fig. 15, Fig. 16, and Fig. 13. Finally, qualitative examples of the detector's output on test scenes are provided in Fig. 18.

## 11. Baseline Model Details

### 11.1. BEV-Based Detectors

PillarNet [9]. A lightweight BEV detector that partitions the point cloud into vertical pillars and applies a 2D CNN backbone for fast inference.

PointNet++ (BEV projection) [28]. We adopt PointNet++ as a BEV feature extractor by projecting point-wise

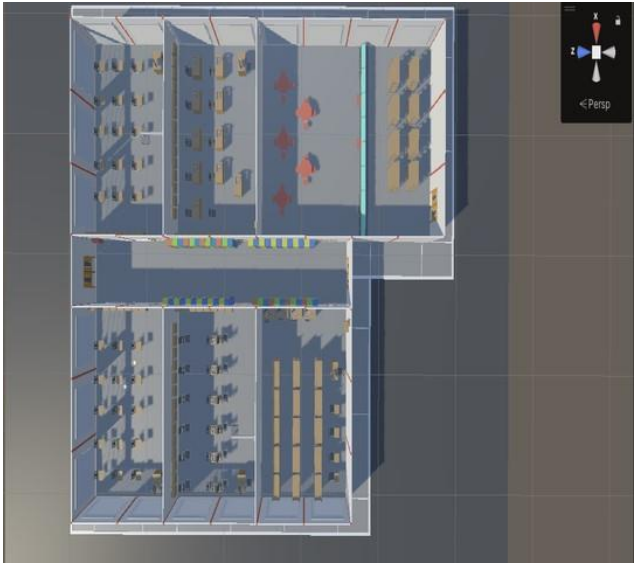
(a) Classroom environment with structured desk arrangements and service workstations.

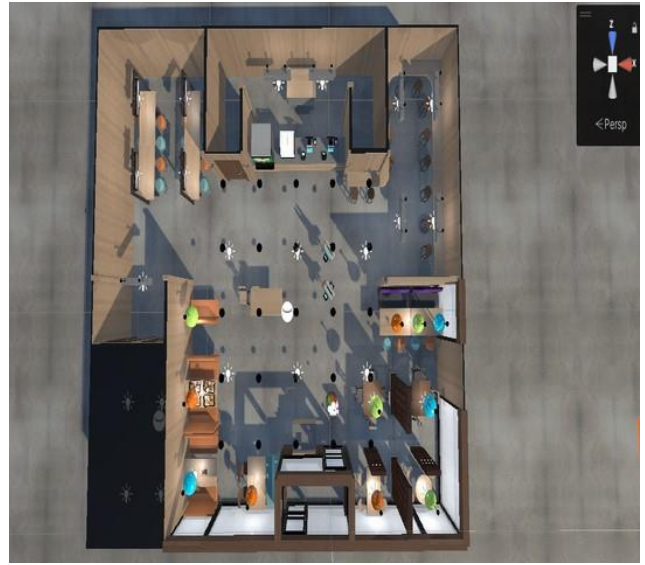
(b) Restaurant setting with varied dining furniture arrangements and areas.

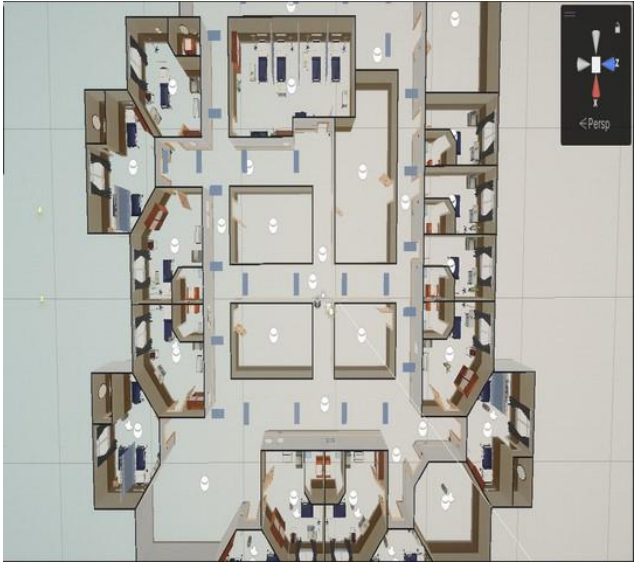
(c) Hospital environment with patient rooms surrounding central corridors, sensor collection points marked.

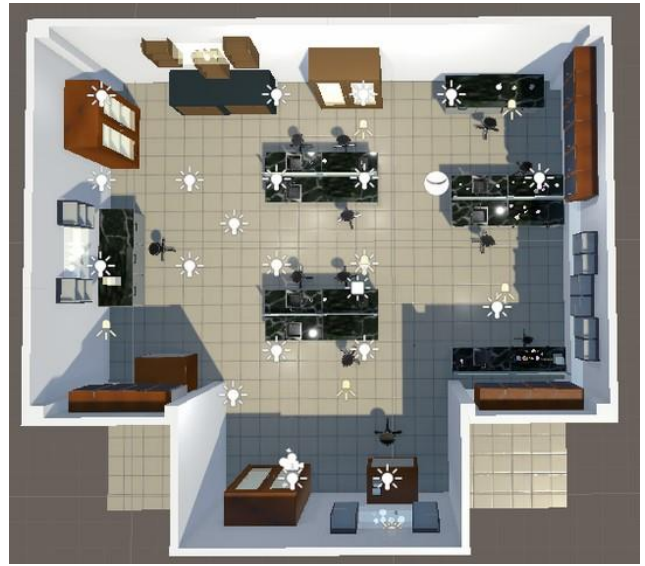
(d) Laboratory workspace with specialized equipment and workstations.

Figure 8. Representative synthetic environments from the INDOOR-LIDAR dataset showcasing diverse indoor layouts: (a) classroom, (b) restaurant, (c) hospital, and (d) laboratory. Each environment features different architectural configurations, furniture arrangements, and complexity levels for comprehensive evaluation of 3D perception algorithms.

features into a discrete BEV grid.

**BEV MAE** [21]. A self-supervised masked autoencoder trained to reconstruct BEV patches, producing robust representations beneficial for low-signal indoor LiDAR data.

**GroupFree3D-BEV** [22]. A transformer architecture adapted for BEV by applying grouping and attention operations on BEV patches rather than point clusters.

### 11.2. Full 3D Object Detectors

**VoxelNet** [60]. A pioneering voxel-based model that learns point features within voxels and aggregates them using 3D convolutions.

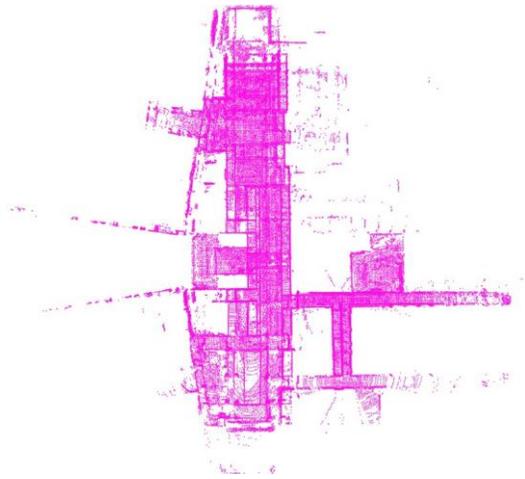

Figure 9. LiDAR point cloud representation of a real-world indoor environment captured for the INDOOR-LIDAR dataset.

SECOND [54]. An improved voxel detector using submanifold sparse 3D convolutions for significantly faster inference.

PointRCNN [37]. A two-stage, point-based architecture that generates proposals directly from raw points and refines bounding boxes using point-level features.

VoteNet [29]. A cornerstone indoor detector leveraging Hough voting on learned point features to predict object centers, widely used in indoor datasets like ScanNet/Matterport3D.

GroupFree3D [22]. A transformer-based point detector that replaces hand-designed voting or anchors with learned grouping and self-attention, achieving state-of-the-art indoor 3D detection accuracy.

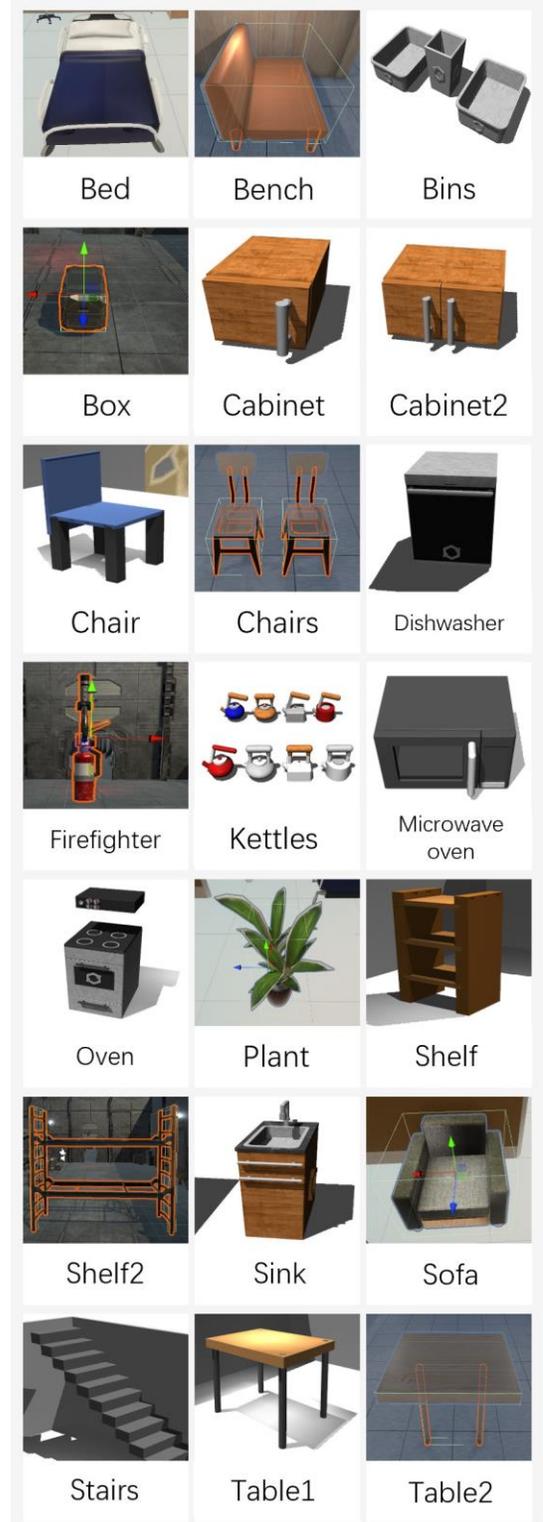

Figure 10. A selection of 3D object models from our asset library used to populate the simulation environment. The collection features a variety of common indoor items, including furniture, appliances, and storage, to support the training and testing of perception pipelines.

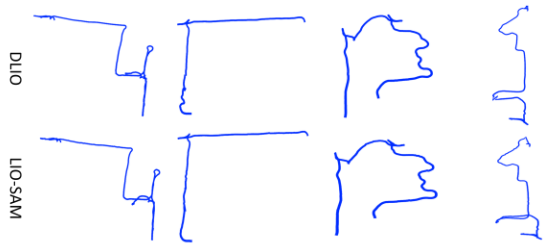

Figure 11. Qualitative comparison of SLAM trajectories from our real-world dataset. The results from DLIO (top row) and LIOSAM (bottom row) are shown for several different paths. The close similarity between the outputs highlights the dataset's quality and its utility for robust SLAM evaluation.

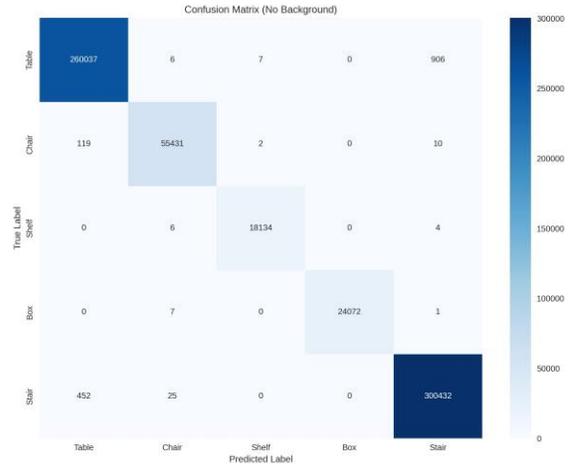

Figure 14. Confusion matrix for the BEV detector's classification task on the simulation test set, showing inter-class misclassification rates.

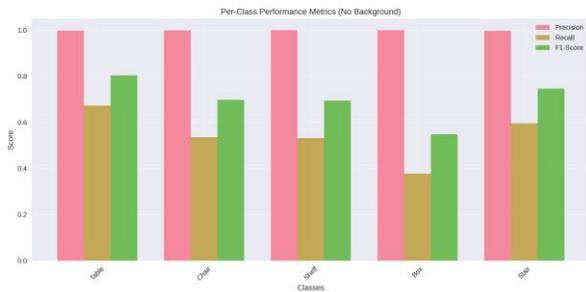

Figure 12. Detailed per-class classification performance of the BEV detector on the simulation test set, showing Precision, Recall, and F1-scores for each object category.

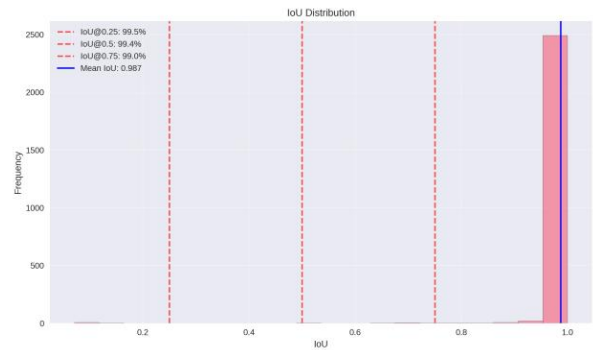

Figure 15. Distribution of Intersection over Union (IoU) scores for all correct detections on the simulation test set. The high concentration at the right (IoU > 0.9) aligns with the 0.987 Mean IoU reported.

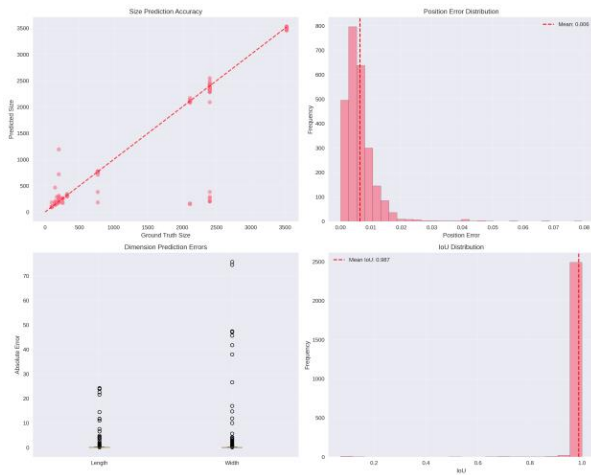

Figure 13. Detailed analysis of bounding box regression performance, visualizing key error metrics (e.g., L1/L2 loss) for the simulation test set.

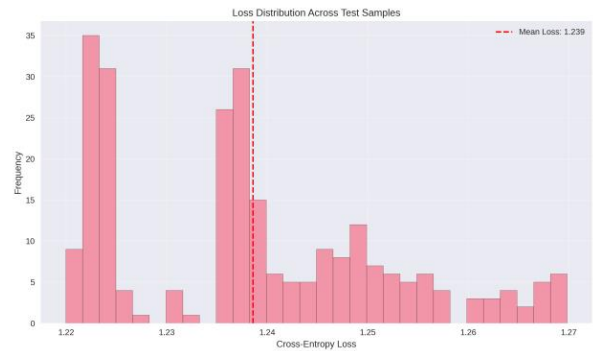

Figure 16. Distribution of the bounding box regression loss for test set samples, indicating that most predictions have very low error.

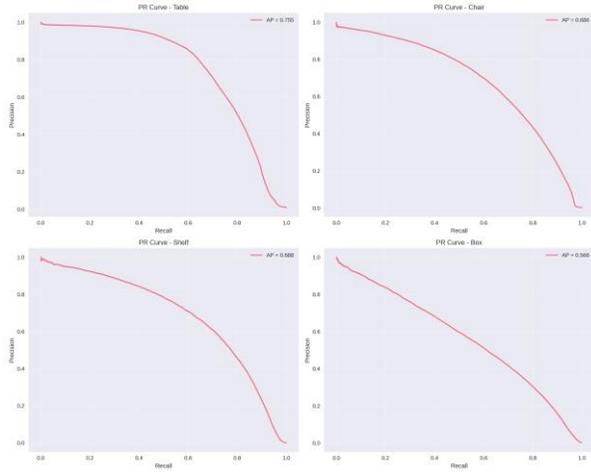

Figure 17. Precision-Recall (PR) curves for the BEV object detector on the simulation test set. The Mean Average Precision (mAP) of 0.679 is computed from these curves.

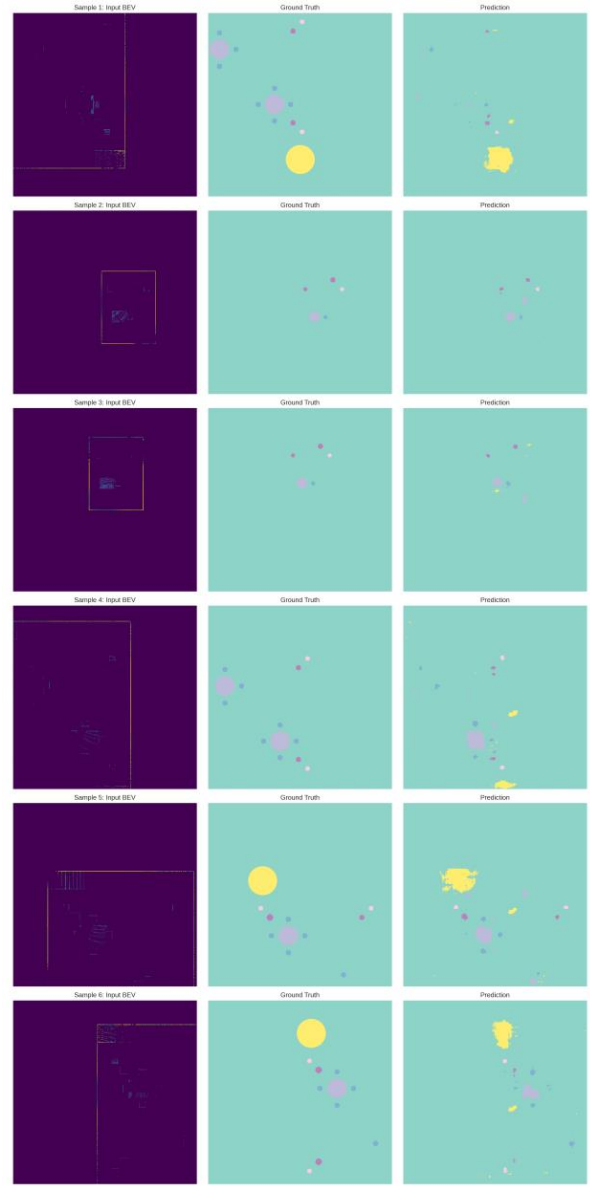

Figure 18. Qualitative examples of the BEV detector's predictions on sample scenes from the simulation test set. This visualization shows the predicted bounding boxes and class labels overlaid on the input point cloud.

## References

[1] Gilad Baruch, Zhuoyuan Chen, Afshin Dehghan, Tal Dimry, Yuri Feigin, Peter Fu, Thomas Gebauer, Brandon Joffe, Daniel Kurz, Arik Schwartz, and Elad Shulman. Arkitscenes: A diverse real-world dataset for 3d indoor scene understanding using mobile rgb-d data, 2022. 3

[2] Nico Brandes. Light sensitive drug products need protection. *West Pharma Blog*, 2020. 2

[3] Angel Chang, Angela Dai, Thomas Funkhouser, Maciej Halber, Matthias Niessner, Manolis Savva, Shuran Song, Andy Zeng, and Yinda Zhang. Matterport3D: Learning from RGBD data in indoor environments. *International Conference on 3D Vision (3DV)*, 2017. 2, 3

[4] Kenny Chen, Ryan Nemiroff, and Brett T Lopez. Direct lidar-inertial odometry: Lightweight lio with continuoustime motion correction. *arXiv preprint arXiv:2203.03749*, 2022. 7, 1

[5] Angela Dai, Angel X. Chang, Manolis Savva, Maciej Halber, Thomas Funkhouser, and Matthias Nießner. Scannet: Richly-annotated 3d reconstructions of indoor scenes, 2017. 2, 3

[6] Epic Games. Unreal Engine. https://www.unrealengine.com/, 2023. Accessed: 2024-05-01. 4

[7] Maohao Fang, Yihan Wang, Sida Zhou, Yue Wang, and Yong Liu. SN-Student: Unsupervised domain adaptation for 3D object detection. In *2021 IEEE/CVF International Conference on Computer Vision Workshops (ICCVW)*, pages 3170–3179, 2021. This work presents a domain adaptation technique for 3D object detection that leverages both labeled synthetic LiDAR data and unlabeled real-world LiDAR data. The approach demonstrates the viability of transferring knowledge from simulation to reality by exploiting the geometric consistency of LiDAR. 5

[8] Andreas Geiger, Philip Lenz, and Raquel Urtasun. Are we ready for autonomous driving? the kitti vision benchmark suite. In *2012 IEEE Conference on Computer Vision and Pattern Recognition*, pages 3354–3361, 2012. 5

[9] Chao Ma Guangsheng Shi, Ruifeng Li. Pillarnet: Realtime and high-performance pillar-based 3d object detection. *ECCV*, 2022. 7, 1

[10] Ulas Gunes, Matias Turkulainen, Xuqian Ren, Arno Solin, Juho Kannala, and Esa Rahtu. Fiord: A fisheye indooroutdoor dataset with lidar ground truth for 3d scene reconstruction and benchmarking. 2025. 4

[11] Yanwen Guo, Yuanqi Li, Dayong Ren, Xiaohong Zhang, Jiawei Li, Liang Pu, Changfeng Ma, Xiaoyu Zhan, Jie Guo, Mingqiang Wei, Yan Zhang, Piaopiao Yu, Shuangyu Yang, Donghao Ji, Huisheng Ye, Hao Sun, Yansong Liu, Yinuo Chen, Jiaqi Zhu, and Hongyu Liu. Lidar-net: A real-scanned 3d point cloud dataset for indoor scenes. In *2024 IEEE/CVF Conference on Computer Vision and Pattern Recognition (CVPR)*, pages 21989–21999, 2024. 2, 3, 4

[12] Timo Hackel, Nikolay Savinov, Lubor Ladicky, Jan Dirk Wegner, Konrad Schindler, and Marc Pollefeys. Semantic3d.net: A new large-scale point cloud classification benchmark. In *ISPRS Annals of the Photogrammetry, Remote Sensing and Spatial Information Sciences*, pages 91–98, 2017. 3

[13] Wolfgang Hess, Damon Kohler, Holger Rapp, and Daniel Andor. Real-time loop closure in 2D LIDAR SLAM. In *2016 IEEE International Conference on Robotics and Automation (ICRA)*, pages 1271–1278, 2016. This paper, foundational for Cartographer, describes a system built for a backpack platform. While successful, it highlights the importance of comprehensive sensor data for robust SLAM, implying that data gaps from occlusions could compromise performance in autonomous navigation scenarios. 2

[14] Yuanming Hu, Tzu-Mao Li, Luke Anderson, Jonathan Ragan-Kelley, and Fredo Durand. Taichi: a language for high-performance computation on spatially sparse data structures. *ACM Transactions on Graphics (TOG)*, 38(6): 1–16, 2019. 5

[15] Aecheon Jung, Soyun Choi, Junhong Min, and Sungeun Hong. Iam: Enhancing rgb-d instance segmentation with new benchmarks, 2025. 3

[16] Harri Kaartinen, Juha Hyyppa, Xiaowei Yu, Markus Holopainen, Matti Vaaja, and Antero Kukko. Personal laser scanning. In *ISPRS Annals of the Photogrammetry, Remote Sensing and Spatial Information Sciences, Volume I-5*, pages 309–314, 2012. Discusses the development and challenges of personal laser scanning (PLS), implicitly highlighting how the operator's presence affects data collection, leading to incomplete captures compared to static or trolley-based systems. 2

[17] Pushyami Kaveti, Aniket Gupta, Dennis Giaya, Madeline Karp, Colin Keil, Jagatpreet Nir, Zhiyong Zhang, and Hanumant Singh. Challenges of indoor slam: A multi-modal multi-floor dataset for slam evaluation, 2023. 4

[18] N. Koenig and A. Howard. Design and use paradigms for gazebo, an open-source multi-robot simulator. In *2004 IEEE/RSJ International Conference on Intelligent Robots and Systems (IROS) (IEEE Cat. No.04CH37566)*, pages 2149–2154 vol.3, 2004. 4

[19] Anton Koval, Vsevolod Hovorun, Roman Bartosh, and Mariusz Olszewski. Comprehensive Survey on SLAM in Challenging Environments for Autonomous Robots and Vehicles. *Applied Sciences*, 13(9), 2023. 2

[20] E Li, Shuaijun Wang, Chengyang Li, Dachuan Li, Xiangbin Wu, and Qi Hao. Sustech points: A portable 3d point cloud interactive annotation platform system. In *2020 IEEE Intelligent Vehicles Symposium (IV)*, pages 1108–1115, 2020. 6

[21] Zhiwei Lin, Yongtao Wang, Shengxiang Qi, Nan Dong, and Ming-Hsuan Yang. Bev-mae: bird's eye view masked autoencoders for point cloud pre-training in autonomous driving scenarios. In *Proceedings of the Thirty-Eighth AAAI Conference on Artificial Intelligence and Thirty-Sixth Conference on Innovative Applications of Artificial Intelligence and Fourteenth Symposium on Educational Advances in Artificial Intelligence*. AAAI Press, 2024. 7, 2

[22] Ze Liu, Zheng Zhang, Yue Cao, Han Hu, and Xin Tong. Group-free 3d object detection via transformers. *arXiv preprint arXiv:2104.00678*, 2021. 7, 2, 3

[23] Viktor Makoviychuk, Lukasz Wawrzyniak, Yunrong Guo, Michelle Lu, Kier Storey, Miles Macklin, David Hoeller, Nikita Rudin, Arthur Allshire, Ankur Handa, and Gavriel State. Isaac gym: High performance gpu-based physics simulation for robot learning, 2021. 4

[24] Daniel Munoz, J. Andrew Bagnell, Nicolas Vandapel, and Martial Hebert. Contextual classification of 3d point clouds. In *2009 IEEE Workshop on 3D Object Recognition*, pages 1–8, 2009. 3

[25] Maxime Merizette, Nicolas Audebert, Pierre Kervella, and Jérôme Verdun. 3dses: an indoor lidar point cloud segmen-

ˆ tation dataset with real and pseudo-labels from a 3d model, 2025. 2, 3, 4

[26] François Pomerleau, Francis Colas, Roland Siegwart, and Stephane Magnenat. Comparing icp variants on real-world data sets: Full-featured, high-performance point cloud registration. *Autonomous Robots*, 39(1):1–18, 2015. This article provides a comprehensive comparison of point cloud registration algorithms, which are fundamental to LiDAR-based mapping and localization, demonstrating the technology's central role in creating high-fidelity spatial representations. 2

[27] Charles R Qi, Hao Su, Kaichun Mo, and Leonidas J Guibas. Pointnet: Deep learning on point sets for 3d classification and segmentation. *arXiv preprint arXiv:1612.00593*, 2016. 5

[28] Charles R. Qi, Li Yi, Hao Su, and Leonidas J. Guibas. Pointnet++: deep hierarchical feature learning on point sets in a metric space. In *Proceedings of the 31st International Conference on Neural Information Processing Systems*, page 5105–5114, Red Hook, NY, USA, 2017. Curran Associates Inc. 5, 7, 1

[29] Charles R. Qi, Or Litany, Kaiming He, and Leonidas J. Guibas. Deep hough voting for 3d object detection in point clouds, 2019. 5, 7, 3

[30] Ishraq Rached, Rafika Hajji, Tania Landes, and Rashid Haffadi. Structscan3d v1: A first rgb-d dataset for indoor building elements segmentation and bim modeling. *Sensors*, 25 (11), 2025. 3

[31] Santhosh K. Ramakrishnan, Aaron Gokaslan, Erik Wijmans, Oleksandr Maksymets, Alex Clegg, John Turner, Eric Undersander, Wojciech Galuba, Andrew Westbury, Angel X. Chang, Manolis Savva, Yili Zhao, and Dhruv Batra. Habitatmatterport 3d dataset (hm3d): 1000 large-scale 3d environments for embodied ai, 2021. 3

[32] J.D. Ribeiro, R.B. Sousa, J.G. Martins, A.S. Aguiar, F.N. Santos, and H.M. Sobreira. IILABS 3D: iilab Indoor LiDAR-based SLAM Dataset, 2025. [Dataset]. 4

[33] Xavier Roynard, Jean-Emmanuel Deschaud, and François Goulette. Paris-lille-3d: A point cloud dataset for urban scene segmentation and classification. In *2018 IEEE International Conference on Robotics and Automation (ICRA)*, pages 2140–2144, 2018. 3

[34] Karim Saleh, Krystian Kadem, Guy Andre Boy, and Anis Bounsiar. An overview of the sim-to-real gap for the perception of autonomous systems in modern LiDAR simulators. In *2022 IEEE International Conference on Systems, Man, and Cybernetics (SMC)*, pages 894–901, 2022. This paper reviews the challenges of transferring perception models from simulation to reality (sim-to-real), highlighting that LiDAR's geometric data is generally more robust to the domain gap than camera images, which are sensitive to visual factors like texture and lighting. 2, 5

[35] Andres Serna, Beatriz Marcotegui, Francois Goulette, and Jean-Emmanuel Deschaud. Paris-rue-madame: A 3d semantic labeling of a complete urban street. In *2014 IEEE/RSJ International Conference on Intelligent Robots and Systems*, pages 4771–4776, 2014. 3

[36] Tixiao Shan, Brendan Englot, Drew Meyers, Wei Wang, Carlo Ratti, and Daniela Rus. Lio-sam: Tightly-coupled lidar inertial odometry via smoothing and mapping. In *2020 IEEE/RSJ international conference on intelligent robots and systems (IROS)*, pages 5135–5142. IEEE, 2020. 7, 1

[37] Shaoshuai Shi, Xiaogang Wang, and Hongsheng Li. Pointrcnn: 3d object proposal generation and detection from point cloud. In *The IEEE Conference on Computer Vision and Pattern Recognition (CVPR)*, 2019. 7, 3

[38] Nathan Silberman, Derek Hoiem, Pushmeet Kohli, and Rob Fergus. Indoor segmentation and support inference from rgbd images. In *Proceedings of the 12th European Conference on Computer Vision - Volume Part V*, page 746–760, Berlin, Heidelberg, 2012. Springer-Verlag. 3

[39] Shuran Song, Samuel P. Lichtenberg, and Jianxiong Xiao. Sun rgb-d: A rgb-d scene understanding benchmark suite. In *2015 IEEE Conference on Computer Vision and Pattern Recognition (CVPR)*, pages 567–576, 2015. 3

[40] Julian Straub, Thomas Whelan, Lingni Ma, Yufan Chen, Erik Wijmans, Simon Green, Jakob J. Engel, Raul Mur-Artal, Carl Ren, Shobhit Verma, Anton Clarkson, Mingfei Yan, Brian Budge, Yajie Yan, Xiaqing Pan, June Yon, Yuyang Zou, Kimberly Leon, Nigel Carter, Jesus Briales, Tyler Gillingham, Elias Mueggler, Luis Pesqueira, Manolis Savva, Dhruv Batra, Hauke M. Strasdat, Renzo De Nardi, Michael Goesele, Steven Lovegrove, and Richard Newcombe. The replica dataset: A digital replica of indoor spaces, 2019. 2, 3 [41] Jiaqi Tan, Weijie Lin, Angel X. Chang, and Manolis Savva. Mirror3d: Depth refinement for mirror surfaces, 2021. 3

[42] Weixuan Tan, Yulan Guo, and Hui-Liang Shen. Toronto-3d: A large-scale mobile lidar dataset for semantic segmentation of urban roadways. In *2020 IEEE/CVF Conference on Computer Vision and Pattern Recognition Workshops (CVPRW)*, pages 2016–2025, 2020. 3

[43] Longbiao Tang, Yongbo Lu, Jia Yu, Yuanxiu Hu, Jingwei Wu, and Chengqun Yu. Backpack-Lidar and Camera-Based Visual and Lidar SLAM for Forests Environments. *Forests*, 14(8), 2023. 2

[44] Sebastian Thrun, Wolfram Burgard, and Dieter Fox. *Probabilistic Robotics*. MIT Press, 2005. A foundational text in robotics that extensively covers how LiDAR is used for key tasks such as mapping, localization, and SLAM (Simultaneous Localization and Mapping), establishing its indispensability. 2

[45] Emanuel Todorov, Tom Erez, and Yuval Tassa. Mujoco: A physics engine for model-based control. *2012 IEEE/RSJ international conference on intelligent robots and systems*, pages 5026–5033, 2012. 4

[46] Maciej Trzeciak, Kacper Pluta, Yasmin Fathy, Lucio Alcalde, Stanley Chee, Antony Bromley, Ioannis Brilakis, and Pierre Alliez. Conslam: Periodically collected real-world construction dataset for slam and progress monitoring. In *Computer Vision–ECCV 2022 Workshops: Tel Aviv, Israel,*

*October 23–27, 2022, Proceedings, Part VII*, pages 317–331. Springer, 2023. 4

[47] Unity Technologies. Unity Real-Time Development Platform. https://unity.com/, 2023. Accessed: 202405-01. 4

[48] Bruno Vallet, Mathieu Bredif, Andres Serna, Beatriz Marcotegui, and Nicolas Paparoditis. Terramobilita/iqmulus urban point cloud analysis benchmark. In *Computers & Graphics*, pages 126–133. Elsevier, 2015. 3

[49] Puthucode Ganesh Vignesh, Nithin Kumar S, Senthil Kumar T, and Jeyakumar G. Lidar-based elderly fall detection system for indoor environments using neural network algorithms. In *2025 4th International Conference on Sentiment Analysis and Deep Learning (ICSADL)*, pages 1564–1568, 2025. 4

[50] Hexiang Wei, Jianhao Jiao, Xiangcheng Hu, Jingwen Yu, Xupeng Xie, Jin Wu, Yilong Zhu, Yuxuan Liu, Lujia Wang, and Ming Liu. Fusionportablev2: A unified multi-sensor dataset for generalized slam across diverse platforms and scalable environments, 2024. 4

[51] Weina Wen, Wenting Dai, Shengxi Fan, Xiyu Zhang, Tofael Ahamed Eite, and Sisi Zlatanova. A Review of SLAMBased Indoor Mobile-Mapping Systems: Accuracy, Performance, and Usability. *Remote Sensing*, 12(22), 2020. 2

[52] Benjamin H Williams, Amy Lau, Christopher Martius, Mischa Ruttimann, and Ulrike Grote. A review of the application of terrestrial laser scanning for forested environments. *Forests*, 12(8):1006, 2021. 2

[53] Bicheng Wu, Xuanyu Xu, Jialong Dai, Ang Wan, Peizhen Zhang, Zoe Wei, and Kurt Keutzer. SqueezeSegV2: Improved model structure and unsupervised domain adaptation for road-object segmentation from a LiDAR point cloud. In *The IEEE International Conference on Computer Vision (ICCV)*, 2019. This paper successfully applies unsupervised domain adaptation to a LiDAR segmentation task, using synthetic data for training. It explicitly addresses the domain shift between synthetic and real LiDAR data, reinforcing LiDAR's suitability as a modality for such hybrid approaches. 5

[54] Yan Yan, Yuxing Mao, and Bo Li. Second: Sparsely embedded convolutional detection. *Sensors*, 18(10), 2018. 7, 3

[55] Chandan Yeshwanth, Yueh-Cheng Liu, Matthias Nießner, and Angela Dai. Scannet++: A high-fidelity dataset of 3d indoor scenes. In *Proceedings of the International Conference on Computer Vision (ICCV)*, 2023. 3

[56] Lintong Zhang, Michael Helmberger, Lanke Frank Tarimo Fu, David Wisth, Marco Camurri, Davide Scaramuzza, and Maurice Fallon. Hilti-oxford dataset: A millimeter-accurate benchmark for simultaneous localization and mapping. *IEEE Robotics and Automation Letters*, 8(1):408–415, 2023. 4

[57] Wenshan Zhao, Zhide Qu, Yiwei Zhou, Jingda Zhang, Keke You, and Volker Tresp. Sim-to-real: A survey. *arXiv preprint arXiv:2003.01369*, 2020. This survey provides a comprehensive overview of the sim-to-real problem, defining it as the challenge of transferring policies or models trained in simulation to the real world due to discrepancies between the two domains, particularly in sensory data. 5

[58] Xiting Zhao and Soren Schwertfeger. ¨ 3dref: 3d dataset and benchmark for reflection detection in rgb and lidar data. *3DV*, 2024. 4

[59] Wenxu Zhou, Kaixuan Nie, Hang Du, Dong Yin, Wei Huang, Siqiang Guo, Xiaobo Zhang, and Pengbo Hu. Il3d: A largescale indoor layout dataset for llm-driven 3d scene generation, 2025. 3

[60] Yin Zhou and Oncel Tuzel. Voxelnet: End-to-end learning for point cloud based 3d object detection, 2017. 7, 2